\documentclass[11pt,a4paper]{article}
\usepackage[hyperref]{eacl2021}
\usepackage{times}
\usepackage{latexsym}
\usepackage{tikz-qtree}

\usepackage{graphicx}
\graphicspath{ {./images/} }

\usepackage{microtype}

\aclfinalcopy % Uncomment this line for the final submission
\title{SCoT: Sense Clustering over Time --\\ a tool for analysing lexical change}

\author{Christian Haase{$^\dag$}, Saba Anwar{$^\dag$}, Seid Muhie Yimam{$^\dag$},  Alexander Friedrich{$^\star$}, Chris Biemann{$^\dag$}\\
  {$^\dag$} Language Technology group, Universit{\"a}t Hamburg, Germany \\
  {$^\star$} Institute for Philosophy, TU Darmstadt, Germany \\
  \texttt{\{anwar,yimam,biemann\}@informatik.uni-hamburg.de} \\
  \texttt{haase.mail@web.de} \\
  \texttt{friedrich@phil.tu-darmstadt.de} \\
  \textit{Update of https://aclanthology.org/2021.eacl-demos.23/}
\\}

\date{}

\begin{document}
\maketitle
\begin{abstract}
We present Sense Clustering over Time (SCoT), a novel network-based tool for analysing lexical change. SCoT represents the meanings of a word as clusters of similar words. It visualises their formation, change, and demise. There are two main approaches to the exploration of dynamic networks: the discrete one compares a series of clustered graphs from separate points in time. The continuous one analyses the changes of one dynamic network over a time-span. SCoT offers a new hybrid solution. First, it aggregates time-stamped documents into intervals and calculates one sense graph per discrete interval. Then, it merges the static graphs to a new type of dynamic semantic neighbourhood graph over time. The resulting sense clusters offer uniquely detailed insights into lexical change over continuous intervals with model transparency and provenance.  SCoT has been successfully used in a European study on the changing meaning of `crisis'. 
\end{abstract}
\begin{figure*}[ht]
\centering
\includegraphics[width=0.95\textwidth]{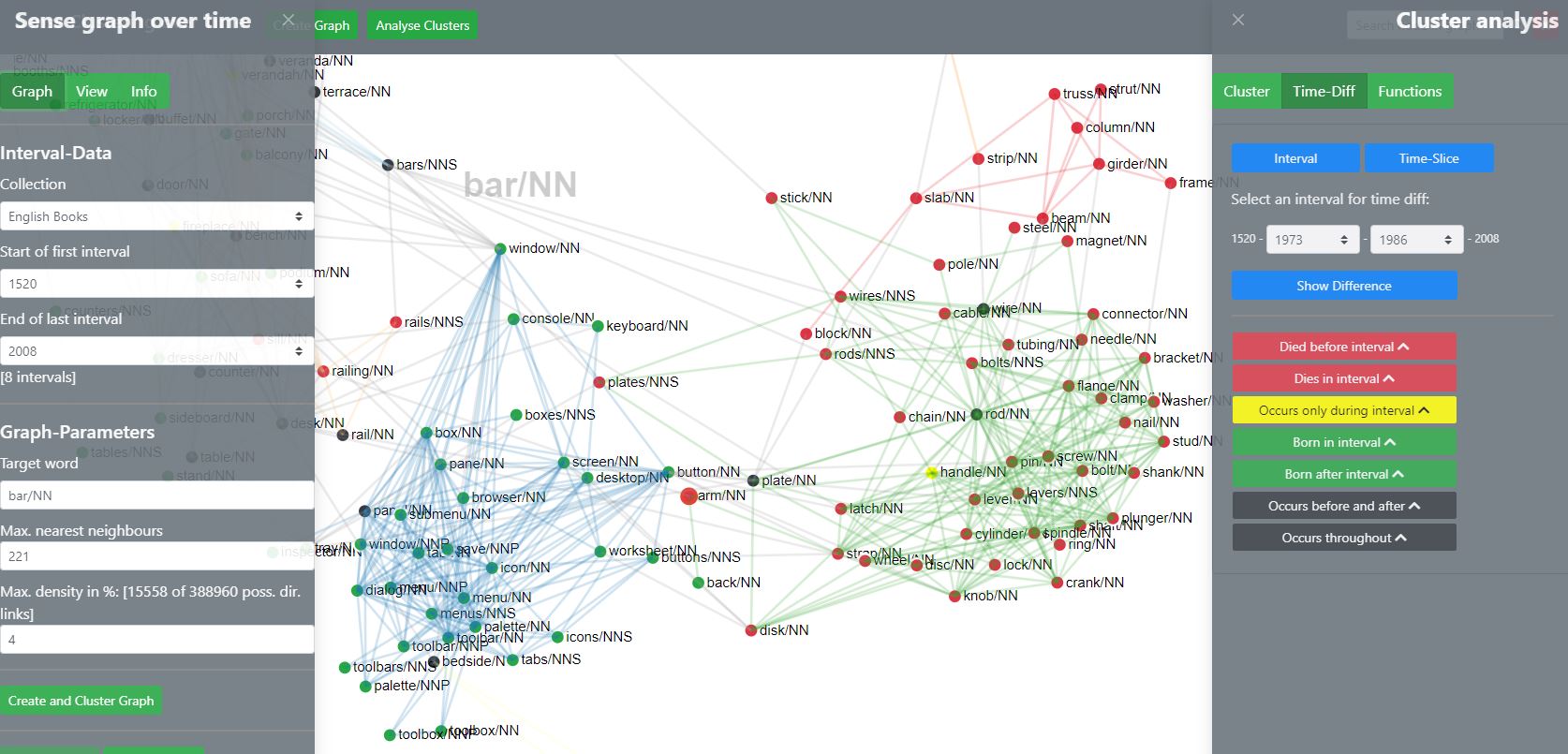}
\caption{\label{fig:scot} Analysis of the sense shifts of `bar/NN' in Google Books \citep{Goldberg:2013} with SCoT: the clusters of the neighbourhood graph over time show that the sense ``a rigid piece of metal used as a fastening or obstruction" [top right] loses traction, while the sense "computer-menu" [bottom left] gains significance. The coloring is relative to the interval "1973-1986". Red indicates the disappearance of a node before 1986. Green indicates the emergence of a node after 1986.}
\end{figure*}
\section{Introduction}
Most real-world networks change over time. So do dynamic networks of word similarities that can be used to infer the meanings of a word. The noun `crisis', for example, used to be strongly linked to the religious word `doom' in English-language books in the early modern period. However, in the modern age `crisis' has become more closely associated with terms denoting economic problems such as `unemployment', `depression' or `inflation' \citep{Biemann:2020}.
\newline
In the recent decade, the interest in dynamic networks has increased. \citep{Rosetti:2018}. This has also stimulated new graph-based approaches to analysing vocabulary change \citep{Mitra:2015, Riedl:2014}. Such research is a key interest of linguists \citep{Tahmasebi:2018, Nulty:2017} and scholars in the humanities \citep{Koselleck:1989, Mueller:2016, Friedrich:2016}. 
\newline
Traditionally, scholars have determined such changes through close reading. However, the growing availability of ever larger digital corpora \citep{Goldberg:2013} and the increasing speed of sense changes in social media \citep{Stilo:2017} have boosted the significance of new research \citep{Nulty:2017}.
\newline
Of particular importance in the research on lexical change is the unsupervised approach of word sense induction (WSI). WSI enables the development of data-driven hypotheses. The approach induces meaning from the bottom upwards and can be used with a diachronic angle. Several implementations for diachronic WSI exist \citep{Tahmasebi:2018}. While many of them represent word meaning by dense vector embeddings, sparse models with network representations still play an important role. The use of sparse, human-readable models enables a better interpretation of meaning hypotheses by linguists and other researchers.
 \newline
There are two main approaches to implement diachronic network-based WSI. Discrete approaches compare several networks that relate to discrete points in time. Continuous approaches analyse when specific nodes, edges or clusters appear in a single dynamic network that changes continuously over time \citep{Rosetti:2018}. 
\newline
Many applications in the field of diachronic network-based WSI fall into the discrete category. \citet{Mitra:2015}, for example, reduce the number of measuring points to single intervals, build one  graph per discrete interval, cluster it and track the resulting sense clusters over intervals. While this approach is considered as less complex than the continuous one \citep{Rosetti:2018}, it brings up complexity problems of its own. The clustering of graphs can namely lead to different solutions. Thus, the number of clustering combinations in a time-series of sense graphs can grow unpredictably. Another issue is the identification of corresponding clusters across time points. 
\newline
Continuous representations are more fine-grained, but can lead to other issues. Since the clustering in such scenarios is mostly done in an incremental way, problems of costly reclusterings or very large clusters can emerge.
\newline
The application Sense Clustering over Time (SCoT) offers a new hybrid approach to network-based WSI that reduces complexity.  SCoT works in two steps. In a first, `discrete' step, the time-stamped documents are aggregated into intervals. Static graphs are built per interval. Then SCoT merges the static graphs to a new type of dynamic neighbourhood graph over time (NGoT). The encoded time-based information from the underlying continuous series of graphs enables a time-coloring of the sense clusters.
\newline
\citet{Haase:2020} has shown that there are different approaches to constructing such semantic NGoTs. The best known method for creating such a dynamic network consists of the merging of a series of equally-sized graphs from each interval, but it is also possible to aggregate nodes and links in different ways. These approaches exhibit different strengths and are explained in more detail below.
\newline
Figure \ref{fig:scot} shows how such a NGoT looks like. The graph shows sense clusters for the target word \textit{bar}. It shows that words such as ``button", ``desktop" and ``icon" became increasingly similar to each other and to the target \textit{bar} in the 1990s, thereby forming the new sense of `computer-menu'.
\newline
SCoT can be used for various tasks such as linguistic studies of polysemic words or research into the history of concepts, but also offers a general and new solution to the analysis of dynamic networks with the neighbourhood graph over time. 
\begin{figure*}[ht]
\centering
\includegraphics[width=0.95\textwidth]{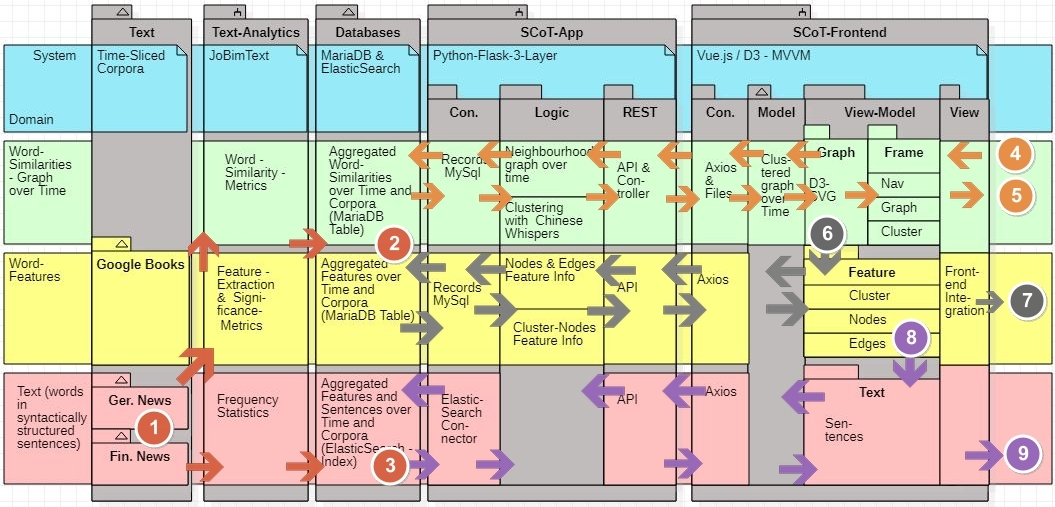}
\caption{\label{fig:architecture} The SCoT-system consists of multiple layers and domain-specific components  relating to the similarity graph (green), the underlying syntagmatic features (yellow) and the corpus (red). The system's four key processes are highlighted with numbers: the preprocessing (1-3), the graph-analysis (4-5), the feature-analysis (6-7) and the querying of example sentences for research (8-9). Arrows indicate the data-flow direction. }
\end{figure*}
\section{System description}
The system enables an analysis of the lexical change of words in an interactive web-interface based on metrics calculated from large time-sliced corpora. This requires a division of the system into a web-front-end and a back-end that accesses the databases with the similarity scores. In addition, the system offers the user additional information on the syntagmatic features that have been used to calculate the similarity scores.
\subsection{Computing distributional thesauri for time-sliced corpora}
The calculation of the similarity scores that inform the graph is steeped in the de Saussurian notion of paradigms and syntagmatic contexts as implemented in JoBimText  \citep{Biemann:2013}. The nodes represent the paradigms. The more syntagmatic contexts two paradigms share, the more similar they are \cite{MC91}. The contexts and words are extracted from the sentences of documents. 
\newline
The raw texts, the syntagmatic features and the network representation of the relations between the paradigms require very different processing steps. They are thus handled by different groups of components of the SCoT application that constitute so-called sub-domains. They have been color-coded in Figure \ref{fig:architecture}.
\newline
The current online demonstration version of SCoT uses three corpora. These include a large dataset of syntactic n-grams from Google Books \citep{Goldberg:2013}, a corpus of Finnish magazines and newspaper articles \citep{lehdet90ff-dl-v2_en}, and a corpus of German newspapers articles as described by \citep{Biemann_theleipzig,  Benikova14networkof}. These corpora have been sliced into 7 to 9 time-based intervals which roughly contain the same amount of data. 
 \newline
The semantic similarity of words can be computed with different methods. For SCoT, we have opted for distributional thesauri (DT) due to their flexibility. They can be based upon different types of context features such as  word n-grams, part-of-speech n-grams and syntactic dependencies. We have used syntactic dependencies.
\newline
We have calculated the DTs with the software JoBimText \citep{Biemann:2013}. It uses the Lexicographer's Mutual Information (LMI) to rank words and their context features. We have limited the computation to the top ranked 1000 features.
\newline
We have stored the scores in one SQL-database per corpus. Each database includes three tables: a table of intervals, a table of word pairs with their similarity scores and references to the intervals in which they occurred, and a table of words and their features with interval information. In addition, we have stored example sentences in an ElasticSearch server. This calculation and storage is highlighted with the numbers 1, 2 and 3 in Figure \ref{fig:architecture}. 
\subsection{Creating the neighbourhood-graph over time}
The system offers the user the possibility to select a target word and to enter parameters for building the NGoT. This is highlighted as number 4 in Figure \ref{fig:architecture}. 
\newline
The user can select between three different types of NGoTs. These have repercussions for the resulting sense clusters \citep{Haase:2020}. We have implemented the interval-, the dynamic and a mixed global/dynamic approach. The user can fine-tune them with three parameters: the number of intervals $i$, the number of nearest neighbour nodes $n$ and the density $d$. 
\newline
The interval-approach creates one static graph with $n$ words and the density $d$ per interval and then merges these. This results in a dynamically sized graph, which is often larger than a static single interval graph. This creates a very clear distinction between clusters and nodes that occur frequently over time and those that do not. The approach is optimal for getting an analytical overview of sense-shifts. We, thus, use it as a starting point for the analysis.
\newline
The dynamic approach fixes the number of unique words $n$ and the density $d$ of the resulting graph and expands the underlying data-points of the static graphs across intervals. This usually only creates partial graphs per interval. Since the dynamic approach fixes the number of links and nodes of the resulting graph it is better suited for comparisons across different graphs than the interval-approach. \newline
The global approach  fixes the number of static single word-nodes and edges in total across all intervals based on maximal values. The significance of the approach lies in the ability to tweak the number of single edges, which has an effect on the number of resulting clusters. For ease of use, we have implemented it as a mixed approach: the nodes are allocated according the dynamic approach. The edges can be tweaked globally.
\newline
The number of the edges in relation to the nodes is the key to creating a useful  graph for clustering and the analysis of lexical change. In order to enhance the dynamic allocation of edges over time, we have relaxed the condition that each node in the resulting graph has a fixed limit of connected edges. This is the standard implementation in many neighbourhood graphs. In sum, SCoT offers a new type of neighbourhood graph that is different to all known implementations of neighbourhood or so-called ego graphs \citep{Mitra:2015}.
\newline
The variants are implemented with a similar pattern: each algorithm first searches for the nodes and then for the edges. Then, the algorithm merges those nodes and edges that refer to the same words in different intervals. It encodes the time-based scores in the nodes and edges.
\subsection{Sense clustering}
The advantage of NGoTs is that they need to be clustered only once. For this, we use the Chinese Whispers algorithm \citep{Biemann:2006}. The key characteristics of the algorithm are that it is non-deterministic, has a linear time-complexity and runs with a fixed number of iterations that result in a stable partition of the graph. We set the number of iterations to 15 in order to increase the chances of the algorithm of reaching a stable plateau. However, there may be more than one stable solution. We have thus implemented the possibility to recluster the graph in order to see whether multiple solutions exist. If one wants to break a tie, it is recommended to slightly reduce the density of the graph and to cluster again. This should remove less significant edges and thus provide a more nuanced clustering.
\subsection{Displaying the sense clusters over time}
During the creation process of the NGoT, the interval information is encoded in the nodes and edges. This information is used for the subsequent coloring of the nodes in the time-difference analysis in the front-end.
\newline
The front-end is based on a modern Model-View-View-Model (MVVM) framework, namely Vue. In MVVM frameworks, the main view of the web-page is rendered by several dynamic model-view components. SCoT has four main components. They render the navigation and side-bars, display the graph, show additional syntagmatic features and exhibit exemplary sentences. The graph-component uses the D3.js library to render the graph. In addition, the front-end includes a connection-layer that communicates with the RESTful SCoT API of the back-end.
\subsection{Diachronic analysis with time-colouring}
Since the sense clusters over time are the most important feature of SCoT and provide the starting point for the research, they are displayed by default when the graph has been created. In the cluster-view, the clusters are ordered by size. 
\newline
The tool offers a wide range of advanced functions to analyse the sense clusters. One can use a hovering function over nodes and links to display the development of similarity scores over time for each node and edge. Furthermore, network metrics such as the betweenness centrality can be used to enlarge central nodes. Such central nodes play a significant role as centres of the clusters and bridges between clusters. Nodes between clusters, which can exhibit ambivalence, can also be highlighted.
\newline
Among the advanced functions, the time-difference mode is particularly noteworthy. The application offers two functions for the time-diff mode.  The first color-codes the nodes in the sense clusters in relation to their occurrence to a set interval. Nodes can disappear before the interval, emerge in the interval or occur after the interval. They can also be stable. The second function offers a slider that highlights all nodes that occur per time-interval \citep{Kempfert:2020}.
\newline
Furthermore, the front-end offers the opportunity to change several view-parameters. These include charge strength, link distance, and the zoom-factor. It is also possible to drag the graph and individual nodes, add name labels to the clusters and manually change cluster assignments. 

\subsection{Model transparency}
A key aim of SCoT is to enable a transparent interpretation of meaning hypotheses. Therefore, SCoT offers functions to drill into the syntagmatic features utilized in the representation of word meaning here. These are available in a count-based sparse model in the form of the DTs from JoBimText \cite{Biemann:2013}. This analysis can be triggered by clicking on a node or an edge. This has been labelled as step 6 in Figure \ref{fig:architecture}. It results in the display syntagmatic contexts per selected word-nodes, including whole clusters, as ranked by LMI. E.g., for the `rod/stick' sense of `bar' in Google books, the most salient syntagmatic contexts are {\small ``-nn/platinum/NN, -dep/stumbling/NN, -dep/altar/JJ, -dep/electro/NN, -nn/vertebral/NN, -in\_pobj/link/NN, -nn/crank/NN, -on\_pobj/leaning/VBG, and\_conj/key/NN"}, whereas the same query for the `menu bar/button' sense yields {\small ``-nn/dialog/NN, -nn/edit/NNP, nn/publishing/NN, -nn/options/NNPS, -on\_pobj/click/NN, -dobj/clicking/VBG, -on\_pobj/button/NN, -nn/cardboard/NN, dep/sill/NN"}.
\newline
The displayed pairs of syntagmatic features and paradigms can serve as a starting point for further analyses: one can retrieve example sentences that include the paradigm and the selected syntagmatic context. This has been labelled as step 8 and 9 in Figure \ref{fig:architecture}.
\begin{figure*}[t]
\includegraphics[width=\textwidth]{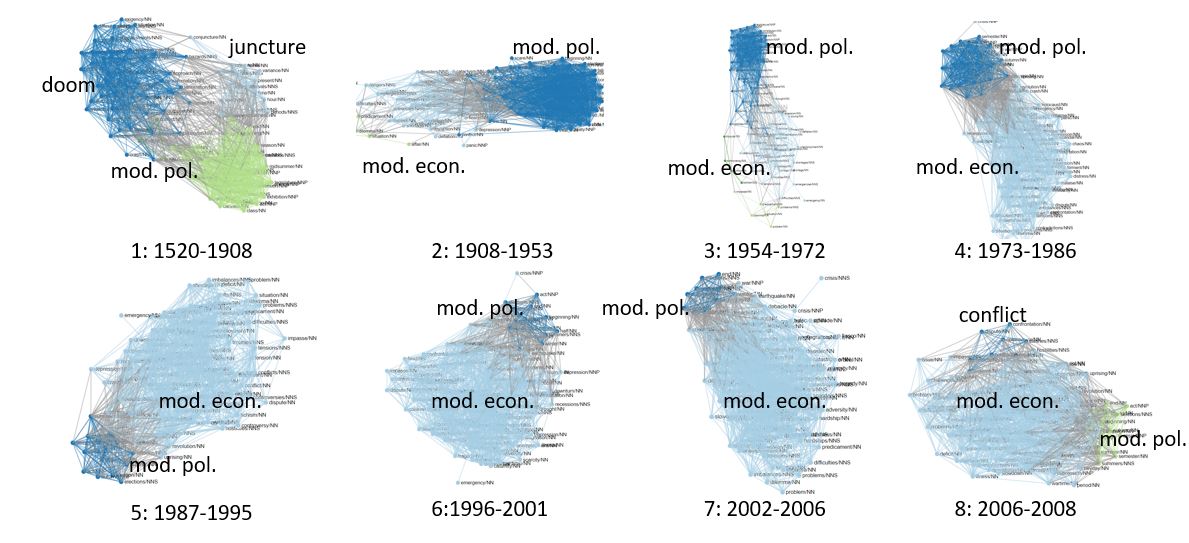}
\caption{Analysis of sense-shifts of crisis/NN: The  neighbourhood graph merges graphs from each interval. The underlying time-series shows that ``crisis/NN" developed a modern political and an economic sense with an increasing dominance of economics between 1520 and 2008. Parameters: n=100, d=30, i=1, corpus: Google Books (English).}
\end{figure*}
\section{Use case: sense shifts of ``crisis"}
SCoT can be used in various research fields. Conceptual history is of particular relevance. It is used to produce lexicons of `basic concepts' and thus encompasses aspects of linguistics and historical research. 
\newline
 \citet{Mueller:2016, Friedrich:2016} have shown that the growing research field is in need of new unsupervised methods in order to deal with newly available large digital corpora. The research that was established by Koselleck places a particular emphasis on concepts that have changed in the transition to the modern age between 1750 and 1850. \citep{Olsen:2012} 
\newline
Within this context, the noun `crisis' takes centre-stage as the contemporaries perceived the transition into the modern age as a time of different crises. The analysis of the changing meanings of the noun is thus an ideal test case for the applicability of the tool in this interdisciplinary field.
\subsection{The `economic turn' and the changing concept of `crisis'}
The first step in most text-based research projects is the choice of the corpus. We have chosen English Google Books as a suitable corpus.
\newline
We start the analysis with a generalized overview over all eight intervals with the graph-type-mode `interval'. We set the parameters n=100 and d=20 and render the graph. This results in a  NGoT with 221 nodes. SCoT analyses three sense clusters over time.
\newline
After we have established the overview, we go into the time-diff mode. From the ongoing research, the prominent hypothesis about the changing meaning of `crisis' between 1750 and 1850 has emerged. We test this hypothesis. We switch to the time-diff mode and color the nodes in relation to the interval 1908-1953. The resulting graph shows that one full sense cluster consists only of `red' nodes that all disappeared in the first interval. We have thus found a candidate for a first sense shift. 
\newline
We now follow up the analysis with a more specific look at the nodes. We find that the pre-modern sense of religious ``doom" and ``juncture of time" was replaced by modern political and economic senses of crisis centred on terms such as `election', `law' or `class'. We then look at individual nodes to deepen the analysis. Each node in the clusters provides an important aspect of the development. The node `class', for example, relates to Marxist philosophy that viewed the cyclic nature of capitalist `crises' as the defining characteristic of the modern age. 
\newline
Subsequently, we want to find out which changes occurred within the modern political and economic clusters in the subsequent intervals. With the help of the time-slider-mode and the individual graphs that are depicted in Figure \ref{fig:architecture}, we can show that that the sense transformation of the term `crisis' continued after the breakthrough of the modern age. An ever growing cluster with economic words can be observed. Terms, such as `depression', `boom', `inflation' and `unemployment' dominate the cluster, increasingly so after the 1950s, and in particular after the oil crisis in the 1970s.
\newline
This tallies with the research on the so-called ``economic turn" in the 1950s and beyond. The argument by economic historians such as Nützenadel is that the cornerstone of the Western postwar-order was the diffusion of new economic and democratic thought, centred on the so-called consensus liberalism that was seen as the antidote to the `crisis' of the great depression and the following political chaos. \citep{Haase:2007} SCoT advances these findings by adding new details to them in a transparent and scientific manner.
\newline
The results of SCoT always need to be contextualised within the limits of the underlying corpus. Google Books contains primary and secondary material and has a strong ``thematic" orientation. Since Google Books contains many books from libraries that serve universities, we need to test whether the `economic turn' of the term `crisis' has shown up so dramatically in the data due to the underlying basis of vast specialist economic literature stored in university libraries. 
\newline
In order to check against the possible bias, we use a second corpus, namely German web-news. We find in this corpus a similar development and conclude that the `economic turn' can be regarded as a wider phenomenon in Western countries after the 1950s. We have arrived at this analysis by the research steps of generalisation, specialisation and comparison that are well supported by SCoT. 

\section{Conclusion and future directions}
This article describes SCoT, a new tool for the analysis of the changes of sense clusters in dynamic networks.  SCoT reduces the complexity of this task through interval-aggregation and neighbourhood graphs over time. The dynamic network retains the time-based information. This enables advanced analyses that can be well visualised. The usage of a sparse approach to distributional semantic modeling provides model transparency and provenance. 
\newline
We have demonstrated the applicability of the solution in the domain of lexical and conceptual change. However, the general nature of the application make it transferable to other domains that use dynamic networks for analysis.
\newline
Future directions in the development of SCoT lie in the further refinement of  neighbourhood graphs over time, the broadening of the usage of SCoT in various domains, including conceptual change, as well the research on the wider implications of the application for diachronic distributional semantics.
\newline
ScoT is available open source under the MIT license\footnote{\url{https://github.com/uhh-lt/SCoT}} and as an online demo\footnote{\url{http://ltdemos.informatik.uni-hamburg.de/scot/}}. A video demonstrating many of the functionalities can be found at \url{https://youtu.be/SbmfA4hKjvg}.
\bibliography{anthology,eacl2021}
\bibliographystyle{acl_natbib}
\end{document}